\documentclass[runningheads]{llncs}
\usepackage[T1]{fontenc}
\usepackage{amsmath,amssymb}
\usepackage{graphicx}
\usepackage{booktabs}
\usepackage{orcidlink}
\usepackage{xcolor,colortbl}

\usepackage{caption}
\usepackage{subcaption}

\usepackage{pgfplotstable}    
\usepackage{pgfplots}
\usepgfplotslibrary{groupplots,dateplot}
\usetikzlibrary{patterns,shapes.arrows}
\pgfplotsset{compat=newest}
\usepgfplotslibrary{groupplots}

\usepackage[nolist]{acronym}

\begin{document}

\newacro{AP}[AP]{Average Precision}
\newacro{cnn}[CNN]{Convolutional Neural Network}
\newacro{dl}[DL]{Deep Learning}
\newacro{dnn}[DNN]{Deep Neural Network}
\newacro{gmm}[GMM]{Gaussian Mixture Model}
\newacro{knn}[kNN]{k-nearest neighbors}
\newacro{mAP}[mAP]{mean Average Precision}
\newacro{mlp}[MLP]{multilayer perceptron}
\newacro{par}[P@$r$]{Precision at $r$}
\newacro{sift}[SIFT]{Scale-Invariant Feature Transform}
\newacro{vit}[ViT]{Vision Transformer}
\newacro{vlad}[VLAD]{Vector of Locally Aggregated Descriptors}
\newacro{wi}[WI]{Writer Identification}
\newacro{wr}[WR]{Writer Retrieval}

\newcommand*{\etal}				{et~al.\;}

\title{Towards the Influence of Text Quantity on Writer Retrieval}
%
%

\author{Marco Peer\orcidlink{0000-0001-6843-0830}\inst{1,2} \and
Robert Sablatnig\orcidlink{0000-0003-4195-1593}\inst{2} \and
Florian Kleber\orcidlink{0000-0001-8351-5066}\inst{2}
}

\authorrunning{M. Peer et al.}

\institute{
iCoSys Institute, University of Applied Sciences and Arts Western Switzerland \\
 \and
Computer Vision Lab, TU Wien, Austria \\
}
\maketitle              
\begin{abstract}
This paper investigates the task of writer retrieval, which identifies documents authored by the same individual within a dataset based on handwriting similarities. While existing datasets and methodologies primarily focus on page level retrieval, we explore the impact of text quantity on writer retrieval performance by evaluating line- and word level retrieval. We examine three state-of-the-art writer retrieval systems, including both handcrafted and deep learning-based approaches, and analyze their performance using varying amounts of text. Our experiments on the CVL and IAM dataset demonstrate that while performance decreases by 20-30\% when only one line of text is used as query and gallery, retrieval accuracy remains above 90\% of full-page performance when at least four lines are included. We further show that text-dependent retrieval can maintain strong performance in low-text scenarios. Our findings also highlight the limitations of handcrafted features in low-text scenarios, with deep learning-based methods like NetVLAD outperforming traditional VLAD encoding.

\end{abstract}
\section{Introduction}
\label{sec:intro}
\begin{figure}[t] \centering \includegraphics[width=0.43\textwidth]{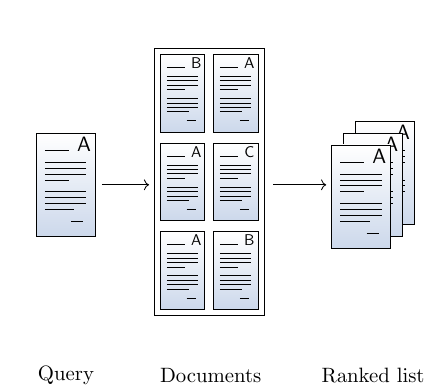} \caption{Definition of \ac{wr}. Documents (Gallery) are ranked based on their similarity to a query, with relevant documents appearing at the top of the ranked list. While \ac{wr} approaches are typically evaluated at the page level, this paper focuses on line- and word level retrieval to investigate the influence of text quantity.}\label{fig:wr_def} \end{figure}

\ac{wr} involves identifying documents written by the same author within a dataset by analyzing handwriting similarities \cite{keglevic_learning_2018}. It is particularly valuable in digital humanities and forensics, helping to identify unknown writers of query documents or track individuals and social groups across historical periods \cite{christlein_icdar_2019}. \ac{wr} operates by using a query document and ranking other documents in the database based on their handwriting similarity, as illustrated in Figure \ref{fig:wr_def}.

Although several datasets exist for both contemporary and historical documents, such as CVL~\cite{cvl}, IAM~\cite{marti_iam-database_2002}, ICDAR2013~\cite{louloudis_icdar_2013}, Historical-WI~\cite{fiel_icdar2017_2017}, and HisIR19~\cite{christlein_icdar_2019}, these datasets and their evaluation benchmarks mainly focus on page level retrieval, where features from an entire page are encoded and aggregated. For contemporary datasets, methods proposed in the literature~\cite{peer_netmvlad,rasoulzadeh} report over 95\% \ac{mAP} for full-page documents. In contrast, this paper addresses \ac{wr} with the question: \textbf{What is the minimum amount of text required for current \ac{wr} methodologies to achieve performance comparable to full-page retrieval}? To our knowledge, no prior research has explored 1) the effect of the amount of text on \ac{wr} and 2) evaluated \ac{wr} at line- or word level granularity. While historical datasets may lack sufficient handwriting samples for certain writers (especially in fragmented sections, such as HisFrag20~\cite{hisfrag20}, but without any provided layout information), contemporary datasets typically include multiple lines per sample. Accordingly, we assess three state-of-the-art \ac{wr} systems (both handcrafted and deep learning-based), consisting of a feature extractor and a codebook-based encoding stage, to evaluate the impact of line- and word level retrieval performance. Our experiments also examine how varying the number of lines per document affects performance, shedding light on the role of text quantity. Additionally, we explore whether current methods can capture writer-specific features on word level, aggregating only features from a single word.

Our results show that performance drops by approximately 20-30\% in terms of \ac{mAP} on both datasets in a short query and gallery setting, where only one text line is used per sample. However, performance remains above 90\% of page level accuracy when at least four lines are included. However, we also demonstrate that retrieval is still effective when the query sample contains limited text and the gallery consists of full pages. Those results also hold when evaluated on Norhandv2~\cite{Tarride2024}, a historical dataset consisting of Norwegian handwriting. For word level \ac{wr}, performance decreases (the best method achieves a \ac{mAP} of about 13\%), but we show that text-dependent retrieval can still identify similarities in these low-text scenarios. Methodologically, we observe that handcrafted features struggle with text scarcity, while NetVLAD outperforms traditional \ac{vlad} encoding in both line- and word level experiments. These findings are valuable for developing and benchmarking future \ac{wr} methodologies, particularly for historical document research where handwriting samples from specific writers are often scarce.

In summary, our contributions are:

\begin{itemize} \item A study on the impact of feature sampling on retrieval performance, \item A detailed line-based evaluation of three state-of-the-art methods to assess the effect of text quantity on retrieval performance, and \item A word-based evaluation of these methods to analyze writing style and word-specific characteristics in extracted features. \end{itemize}

The rest of the paper is organized as follows: Section \ref{sec:relWork} discusses related work in \ac{wr}. Section~\ref{sec:methodology} provides an overview of the state-of-the-art \ac{wr} methods used in this work, along with the datasets for our experiments. The results are presented in Section \ref{sec:evaluation}, followed by the conclusion in Section \ref{sec:conclusion}.

\section{Related Work}
\label{sec:relWork}

In the following, we give an overview of related work for \ac{wr}, followed by a brief summary of similar work on historical documents. 
The first method for \ac{wr} based on deep learning is established by Fiel and Sablatnig \cite{azzopardi_writer_2015} who propose a CaffeNet, a \ac{cnn}, trained with segmented text lines in order to build page descriptors using mean pooling of the activations of the penultimate layer.  In \cite{christlein_cnn_vlad}, Christlein et al. train a \ac{cnn} but aggregate the features with VLAD and triangluation embeddings. Christlein \etal \cite{gall_offline_2015} extract local features from image patches, using a \ac{cnn} architecture trained on patches centered around handwriting contours. The embeddings derived from the second-last layer's activation functions are encoded with GMM supervectors \cite{christlein_writer_2014}, followed by a ZCA-whitening normalization step. In contrast to SIFT or neural networks, Christlein \etal \cite{zernike} perform the retrieval with \ac{vlad}-encoded Contour Zernike moments, a handcrafted local shape descriptor. Keglevic \etal \cite{keglevic_learning_2018} introduce a Triplet \ac{cnn} approach, where image patches centered at \ac{sift} \cite{low-04} key points are used to train a three-branch network based on a similarity measure. The network’s final linear layer serves as an embedding, which is encoded using multiple \ac{vlad}~\cite{arandjelovic_all_2013} encodings of different vocabulary sizes. Unlike most \acs{dl}-based approaches that train \acp{cnn} from scratch, Liang \etal \cite{sun_offline_2021} leverage transfer learning by employing a pre-trained ResNet-50 \cite{he_deep_2016} in combination with \ac{vlad} \cite{arandjelovic_all_2013} for encoding. In \cite{rasoulzadeh}, Rasoulzadeh and Babaali employ NetVLAD~\cite{netvlad}, a learned variant of the \ac{vlad} encoding. Their approach is currently the state of the art on contemporary datasets such as CVL~\cite{cvl} or ICDAR2013~\cite{louloudis_icdar_2013}. Additionally, Peer \etal~\cite{peer_netmvlad} investigate NetMVLAD, an extension with multiple vocabularies within the NetVLAD layer, that slightly outperforms NetVLAD. 

For historical documents, Christlein \etal \cite{unsupervised_icdar17} propose an unsupervised approach with VLAD encodings based on ResNet20 operating on patches extracted at SIFT keypoint locations. Their target labels are generated by clustering the respective SIFT descriptors. This training strategy is also applied by the winners of the competition of the respective dataset called HisFrag20~\cite{hisfrag20}, that contains artificially created fragments of medieval manuscripts. Those samples contain less text, usually up to a few words. Peer \etal~\cite{peer_saghog_2024} show that with self-supervised pretraining, the performance for \ac{wr} is improved when having samples with less text, but the performance scores are lower than for full-page datasets (46.6\% \ac{mAP} without pretraining \cite{avlad,peer_saghog_2024}). Raven et al.\cite{Raven2024} achieve state-of-the-art performance on two historical datasets consisting of full pages by self-supervised training of a transformer-based architecture combined with \ac{vlad}. Mattick et al.~\cite{Mattick2024} investigate the use of learned poolings for the task of \ac{wr} on the Historical-WI~\cite{fiel_icdar2017_2017}, showing that traditional sum pooling is superior.

While the methods described in this section are all text-independent, a first approach towards text-dependent \ac{wr} by applying a character- and $n$-gram-based scheme is introduced by Peer \etal \cite{beer25} in the domain of Greek papyri. They investigate the performance when only aggregating specific characters, such as the trigram \textit{kai} (\textit{and} in Greek). However, this finding is limited to Greek texts, and no general experiments have been conducted regarding the influence of text quantity on historical or contemporary datasets.

\section{Methodology}\label{sec:methodology}

We use three state-of-the-art approaches for \ac{wr}, namely
\begin{itemize}
    \item SIFT + \ac{vlad},
    \item ResNet20 + \ac{vlad} 
    \item and ResNet20 + NetVLAD~\cite{netvlad}.
\end{itemize}

The approaches are chosen to cover different aspects of the respective \ac{wr} pipeline: The feature extraction stage is either based on handcrafted features in case of SIFT or deep-learning-based with ResNet20. Regarding the encoding stage, we rely on an externally trained codebook (\ac{vlad}) or an encoding jointly trained with the feature extractor in an end-to-end manner (NetVLAD). Although all approaches are based on \ac{vlad}, as stated in Section~\ref{sec:relWork}, the methods proposed in related work - in particular those leading the performance scores for \ac{wr}~\cite{zernike,rasoulzadeh} - are based on this encoding.

\subsection{Sampling} 
\begin{figure}[t]
    \centering
    \includegraphics[width=0.75\linewidth]{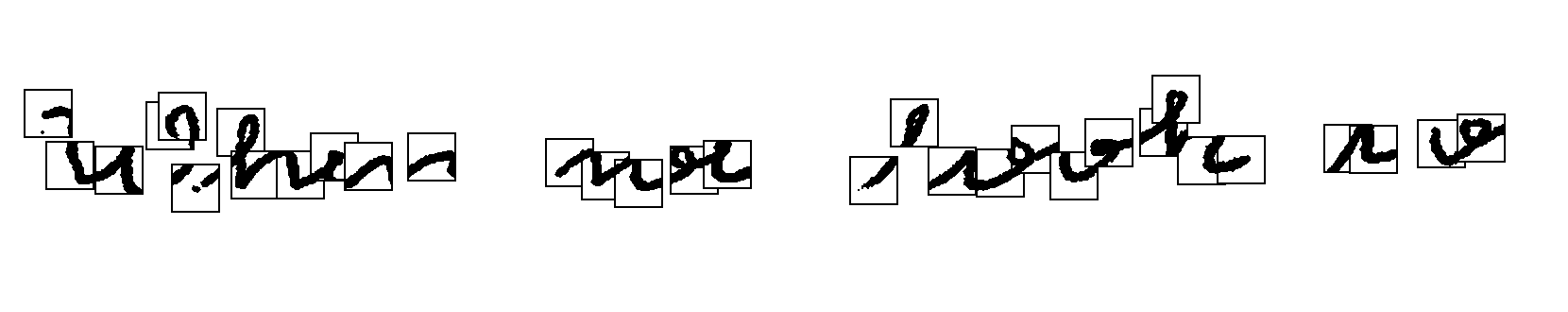}
    \caption{$32\times32$ patch sampling.}
    \label{fig:patch-sampling}
\end{figure}

Instead of relying on a keypoint detection algorithm, usually \ac{sift}~\cite{unsupervised_icdar17,peer_netmvlad,rasoulzadeh}, we directly sample keypoints at the contour of the binarized images (black pixels as they represent the ink) as this improves performance, in particular when handwriting is scarce~\cite{christlein_handwriting_2018}. With those keypoints, we compute the SIFT descriptors as well as extract $32\times32$ patches that are then forwarded to the neural network. An example of the patch sampling is shown in Fig.~\ref{fig:patch-sampling}.  Blank patches - containing no handwriting - are discarded.

\subsection{Feature Extraction}
In the following, the two feature descriptors used in our word are briefly explained. 

\paragraph{SIFT features} For SIFT descriptors, we use the RootSIFT descriptor, a modified version of the standard SIFT descriptor aimed at improving feature matching proposed by Arandjelovic et al.~\cite{arandjelovic_three_2012}. It consists of a $l_1$ normalization, a square root transformation - each element of the L1-normalized vector is transformed by taking the square root - followed by $l_2$ normalization.

\paragraph{CNN-based features} As a deep-learning-based feature extractor, ResNet20~\cite{he_deep_2016} as applied in related work~\cite{unsupervised_icdar17,rasoulzadeh}, is used. The network is trained on the $32\times32$ patches sampled at the contour of the handwriting. The fully-connected layer for classification is cut off and the embeddings are directly trained via the triplet loss~\cite{balntas_pn-net_2016} on semi-hard triplets. As a target, the writer label is used.

\subsection{Encoding}  \ac{wr} is dominated by \ac{vlad}-based encodings, either by the vanilla version \cite{arandjelovic_all_2013} or the learnable version NetVLAD~\cite{netvlad}. \ac{vlad} clusters a vocabulary to obtain $N_c$ clusters $\{ \boldsymbol{\mathrm{c}}_0, \boldsymbol{\mathrm{c}}_1, \dots, \boldsymbol{\mathrm{c}}_{N_c-1}\}$ and encodes a set of local descriptors $\boldsymbol{\mathrm{x}}_i,$ $i \in \{0, \dots, N-1\}$ via

\begin{equation}\label{eq:vlad}
	\boldsymbol{\mathrm{v}}_k =  \sum_{i=0}^{N-1} \boldsymbol{\mathrm{v}}_{k,i} = \sum_{i=0}^{N-1} {\alpha}_k(\boldsymbol{\mathrm{x}}_i)  (\boldsymbol{\mathrm{x}}_i -  \boldsymbol{\mathrm{c}}_k), \quad k \in \{0, \dots, N_c-1\},
\end{equation}

with $\alpha_k = 1$  if $\boldsymbol{\mathrm{c}}_k$ is the nearest cluster center to $\boldsymbol{\mathrm{x}}_i$, otherwise 0, hence making the original VLAD encoding not differentiable. The final global descriptor is then obtained by concatenating the vectors $\boldsymbol{\mathrm{v}}_k$. 

For NetVLAD, Arandjelović et al. \cite{netvlad} suggest the a learnable layer which tackles the non-differentiability of $\alpha_k$ in (\ref{eq:vlad}) by introducing a convolutional layer with parameters $\{\boldsymbol{\mathrm{w}}_{k},b_{k}\}$ for each cluster center $\boldsymbol{\mathrm{c}}_k$ to learn a soft-assignment

\begin{equation}
	\overline{\alpha}_k(\boldsymbol{\mathrm{x}}_i) = \frac{e^{\boldsymbol{\mathrm{w}}_k^\mathrm{T} \boldsymbol{\mathrm{x}}_i + b_k }}{\sum_{k'}^{} e^{\boldsymbol{\mathrm{w}}_{k'}^\mathrm{T} \boldsymbol{\mathrm{x}}_i + b_{k'} }}.
\end{equation}

The cluster centers $\boldsymbol{\mathrm{c}}_k$ of NetVLAD are also learned during training.

\subsection{Global Descriptor and Retrieval} We aggregate all embeddings $\{\boldsymbol{\mathrm{V}}_0, \boldsymbol{\mathrm{V}}_1, \dots, \boldsymbol{\mathrm{V}}_{N_f-1}\}$ of an entity (page, line or level) using $l_2$ normalization followed by sum pooling

\begin{equation}
    \boldsymbol{\mathrm{V}} = \sum_{i=0}^{N_f-1} \boldsymbol{\mathrm{V}}_i
\end{equation}

to get the global descriptor $\boldsymbol{\mathrm{V}}$. Furthermore, to reduce visual burstiness \cite{visual_burstiness}, we apply power-normalization $f(x) = \text{sign}(x)\sqrt{|x|}$, followed by $l_2$-normalization. Finally, a dimensionality reduction with whitening via PCA is performed. For the ranked list, the entities are ranked by calculating the cosine similarity of the global descriptors. 

\subsection{Datasets}\label{subsec:datasets}

Our experiments are conducted on two datasets, the CVL and the IAM database, briefly described in the following, with the statistics shown in Table~\ref{tab:datasets}.

\begin{table}
\centering
    \caption{Statistics of the evaluation datasets used.}\label{tab:datasets}
    \begin{tabular}{lll}
    \toprule
    ~ & CVL  & IAM \\ \midrule
    \# of writers & 283 & 483 \\
    \# of documents & 1409 & 1159\\ 
    \# of lines & 11842 & 8979 \\
    \# of words & 87752 & 72939\\

    \bottomrule
    \end{tabular} 
\end{table}

\paragraph{CVL} The train set of the CVL~\cite{cvl} dataset includes 27 authors who contributed seven texts (one in German and six in English) resulting in 189 pages , whereas the test set consists of 283 authors with five pages (one in German and four in English) each. 
\paragraph{IAM} The IAM~\cite{marti_iam-database_2002} database consists of 1539 pages written by 657 writers. 356 writers contributed only one page while the remaining 301 ones wrote between two and 60 pages. Since there is no official test split, we take all authors contributing two or three pages as a training set (380 pages), resulting in 1159 from 483 writers as our test set.

\begin{figure}
\begin{subfigure}[b]{0.48\textwidth}
    \centering
    \includegraphics[width=\linewidth]{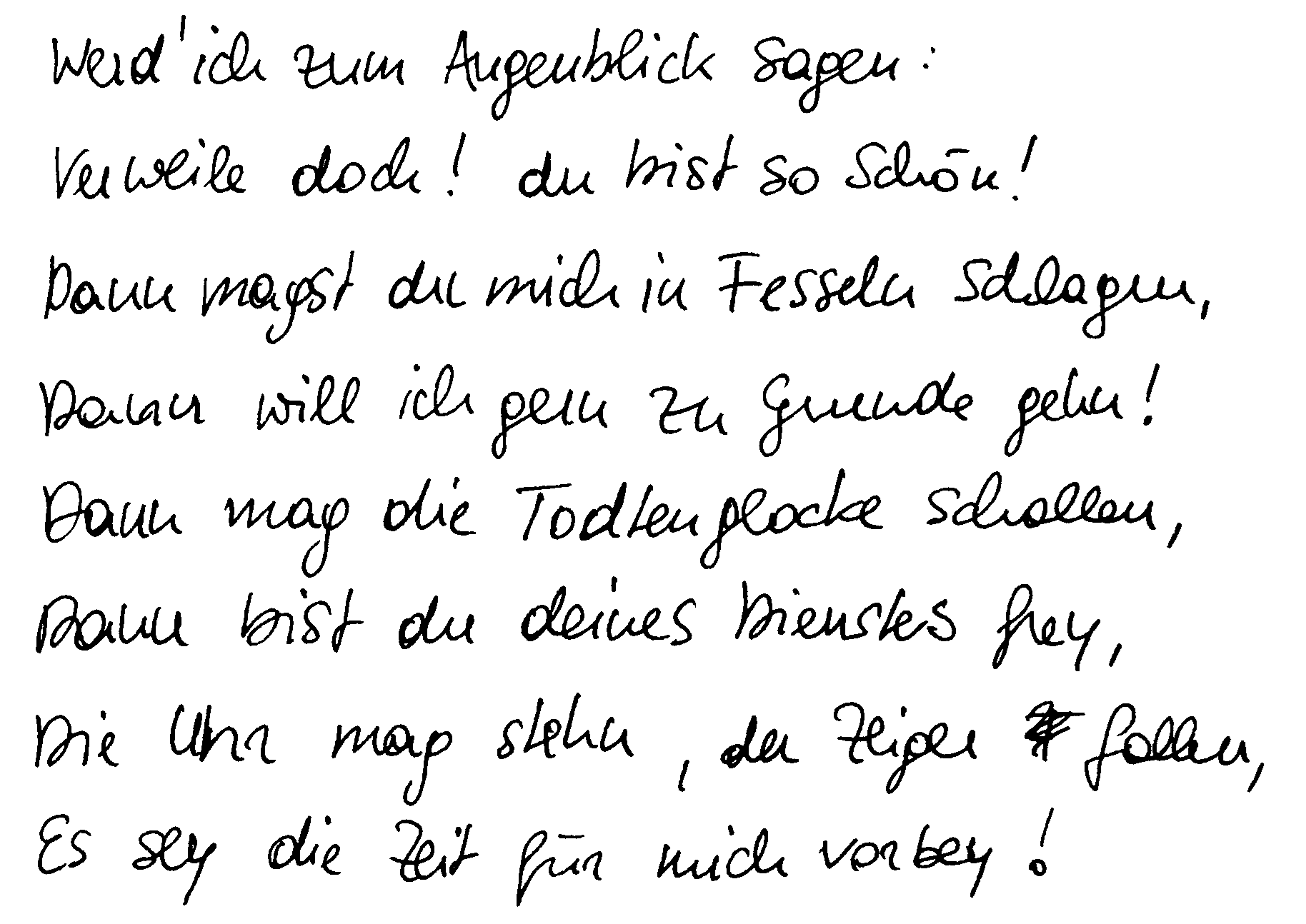}
    \caption{~}
    \label{fig:cvl-sample}
\end{subfigure}\hfill
\begin{subfigure}[b]{0.48\textwidth}
    \centering
    \includegraphics[width=\linewidth]{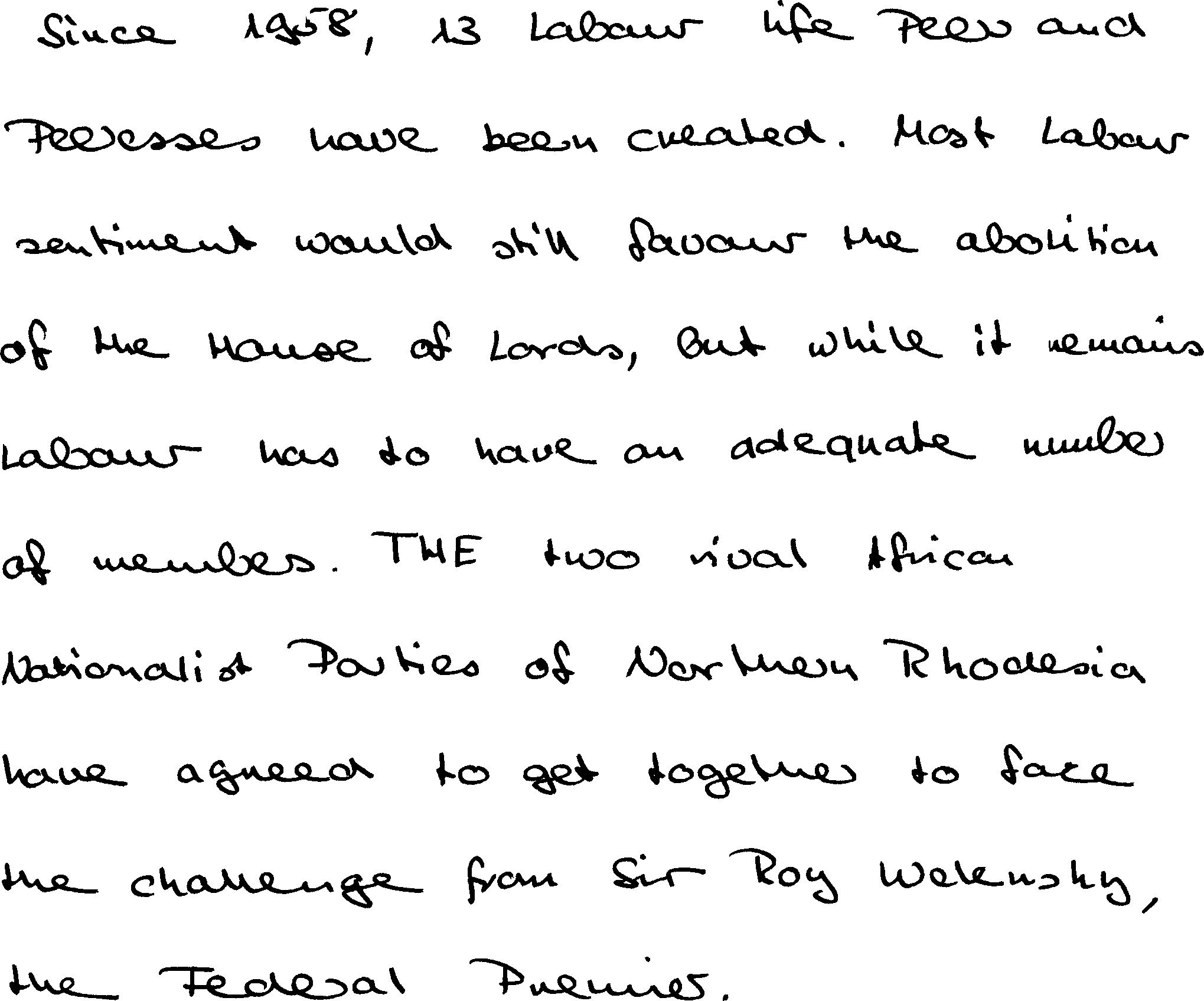}
    \caption{~}
    \label{fig:iam-sample}
\end{subfigure}
\caption{A handwriting sample of the (\subref{fig:cvl-sample}) CVL (ID: 0074-6) and (\subref{fig:iam-sample}) IAM (ID: a01-007u) dataset.}\label{fig:dataset_example}
\end{figure}

We binarize both datasets with Otsu's method \cite{otsu_threshold_1979} and use the provided line and word annotations to perform our experiments. For the IAM dataset, we remove all words only containing special characters. An example of both datasets is shown in Fig.~\ref{fig:dataset_example}.

\section{Results}\label{sec:evaluation}

Next, we present our evaluation protocol followed by the results for \ac{wr} on all three granularities: page-, line- and word level.

\subsection{Evaluation Protocol}

 All methods are evaluated with the same protocol. ResNets are trained with a margin of $m=0.1$ for the triplet loss, the (Net-)\ac{vlad} vocabularies have a fixed size of 100 and the global descriptors are whitened and reduced to the dimension of 256 to ensure a fair comparison. The training is done on the respective training set, as defined in Subsection~\ref{subsec:datasets}. The main metric used for our evaluation is \ac{mAP}: For each $\mathrm{q}$ of the list of queries with length $Q$, the average precision is given by

\begin{equation}
    \mathrm{AP}_\mathrm{q} = \frac{1}{R} \sum_{k=1}^{Q} \mathrm{Pr}_\mathrm{q}(k) \cdot \mathrm{rel}_\mathrm{q}(k),
\end{equation}

where $\mathrm{Pr}_\mathrm{q}(k)$ is the ratio of relevant documents among the first $k$ elements for query $\mathrm{q}$, $R$ is the total number of relevant documents in the gallery and $\mathrm{rel}_\mathrm{q}(k)$ is defined as

\begin{equation}
    \mathrm{rel}_\mathrm{q}(k)= \begin{cases}
1 & \text{if} \ \mathrm{Label}(\mathrm{q}) = \mathrm{Label}(k) \\
0 & \, \text{else}
\end{cases}.
\end{equation}

If not stated otherwise, the label used for evaluation is the writer label. Secondly, we evaluate using the (Hard) Top-$x$ metric, indicating if the first $x$ documents are written by the same writer.

\begin{table}
\centering
    \caption{\ac{wr} performance on page level with 5k features used per line.}\label{tab:page-results}
    \begin{tabular}{p{4cm}cccc}
    \toprule
    ~ & \multicolumn{2}{c}{CVL} & \multicolumn{2}{c}{IAM} \\
    ~ & \ac{mAP} & Top-1 & \ac{mAP} & Top-1 \\ \midrule
    SIFT + \ac{vlad} & 96.1 & 98.3 & 92.5 & 99.4 \\ 
    ResNet20 + \ac{vlad} & \textbf{97.4} & \textbf{99.2} & \textbf{93.8} & \textbf{99.5}\\
    ResNet20 + NetVLAD & 97.0 & 98.9 & 92.2 & 99.4 \\
    \bottomrule
    \end{tabular} 
\end{table}

\subsection{Page level Retrieval}\label{subsec:page}
\begin{figure}
    \centering
        
\definecolor{darkgray176}{RGB}{176,176,176}
\definecolor{steelblue31119180}{RGB}{31,119,180}
\begin{tikzpicture}
\pgfplotsset{grid style={dashed,gray}}
       
\begin{axis}[
legend style={nodes={scale=0.75, transform shape}},
name=ax1,
width=0.48\textwidth,
tick align=inside,
tick pos=left,
xmin=0, xmax=5100,
xtick style={color=black},
ymin=0, ymax=100,
xtick={1000, 2000, 3000, 5000},
xticklabels={1k, 2k, 3k, 5k},
ytick style={color=black},
xlabel= {Number of features},
ylabel=\ac{mAP},
xmajorgrids,
ymajorgrids,
legend style={font=\small, at={(-0.1,0.-0.4)},anchor=north west,legend columns=3},
] 

\addplot [semithick, color=steelblue31119180, mark=*]
table {%
10 4.64
50 12.0
100 20.4
250 39.7
500 61.1
1000 79.4
1500 86.0
2000 88.6
3000 93.4
5000 96.1
};
\addlegendentry{\ac{sift} + \ac{vlad}}

\addplot [semithick, color=orange, mark=*]
table {%
10 30.8
50 70.2
100 83.7
250 92.5
500 95.7
1000 97.0
1500 97.2
2000 97.2
3000 97.5
5000 97.4
};
\addlegendentry{ResNet20 + \ac{vlad}}

\addplot [semithick, color=red, mark=*]
table {%
10 46.0
50 82.7
100 90.7
250 95.3
500 96.4
1000 97.03
1500 96.8
2000 96.8
3000 96.9
5000 97.0
};
\addlegendentry{ResNet20 + NetVLAD}

\legend{}

\end{axis}

\begin{axis}[
legend style={nodes={scale=0.75, transform shape}},
at={(ax1.south east)},
xshift=2cm,
width=0.48\textwidth,
tick align=inside,
tick pos=left,
xmin=0, xmax=5100,
xtick={1000, 2000, 3000, 5000},
xticklabels={1k, 2k, 3k, 5k},
xtick style={color=black},
ymin=0, ymax=100,
ytick style={color=black},
xlabel= {Number of features},
ylabel=\ac{mAP},
xmajorgrids,
ymajorgrids,
legend style={font=\small, at={(-1.2,0.-0.4)},anchor=north west,legend columns=3},
]

\addplot [semithick, color=steelblue31119180, mark=*]
table {%
10 7.4
50 17.1
100 26.2
250 46.0
500 63.0
1000 78.1
1500 84.5
2000 87.0
3000 90.5
5000 92.5
};
\addlegendentry{\ac{sift} + \ac{vlad}}

\addplot [semithick, color=orange, mark=*]
table {%
10 51.7
50 82.0
100 87.7
250 92.0
500 92.9
1000 93.2
1500 93.4
2000 93.6
3000 93.6
5000 93.8
};
\addlegendentry{ResNet20 + \ac{vlad}}

\addplot [semithick, color=red, mark=*]
table {%
10 52.9
50 83.5
100 88.9
250 91.3
500 91.9
1000 91.8
1500 92.0
2000 92.0
3000 92.0
5000 92.2
};
\addlegendentry{ResNet20 + NetVLAD}

\end{axis}
\node at (2.1,3.8) {CVL};
\node at (8.4,3.8) {IAM};

\end{tikzpicture}  
    \caption{\textbf{Page level Retrieval}: Influence of the number of features used per line \ac{wr} performance for CVL and IAM dataset.}\label{fig:page-feats}    
\end{figure}
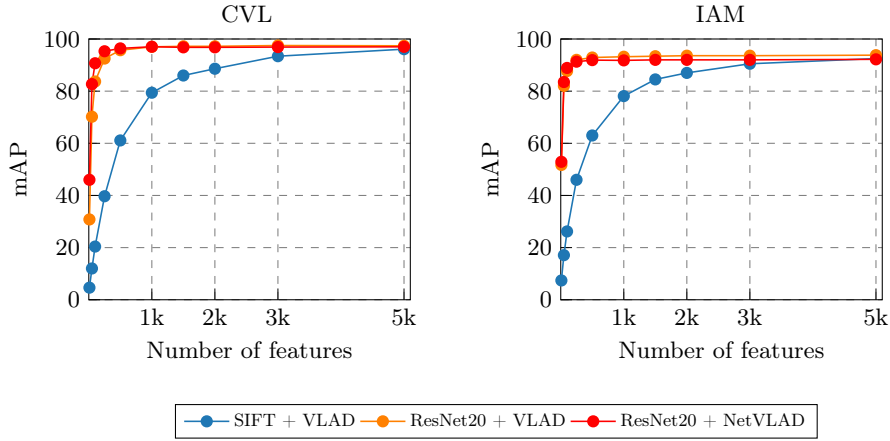

To demonstrate the effectiveness of the selected methods, we evaluate page retrieval, with results reported in Table~\ref{tab:page-results}. The best performing method is ResNet20 with \ac{vlad} on both datasets, though all three methods show similar performance within a 1.3 percentage point \ac{mAP} range. These results are obtained by sampling 5k features per line, yielding approximately 40k features per document. For a fair comparison, we assess the number of features used per line across both datasets in Fig.~\ref{fig:page-feats}. We evaluate a range from 10 to 5000 features (either SIFT descriptors or patch embeddings) and observe that the performance saturates for ResNet-based feature extraction at around 1k features per line, while SIFT continues to improve until 5k, making deep-learning-based approaches more computationally efficient.

\subsection{Line level Retrieval}

The experiments in this section are conducted at the line level, where each line serves as a query and the remaining lines are ranked. Compared to page level \ac{wr}, NetVLAD outperforms \ac{vlad}, as shown in Table~\ref{tab:wr-line} (7.7 and 0.3 percentage points \ac{mAP} for CVL and IAM, respectively). However, a significant performance drop is observed when comparing these results to page level retrieval in Subsection~\ref{subsec:page}, with a decrease of 30/20 percentage points \ac{mAP}. This suggests that all three methods are affected by less text being encoded, with IAM lines performing slightly better than CVL, likely due to IAM lines containing more text on average ($\approx 7.4$ vs. $\approx 8.1$ words per line).

\begin{table}
\centering
    \caption{Line level \ac{wr} with 5000 features used per line.}\label{tab:wr-line}
    \begin{tabular}{p{4cm}cccc}
    \toprule
    ~ & \multicolumn{2}{c}{CVL} & \multicolumn{2}{c}{IAM} \\
    ~ & \ac{mAP} & Top-1 & \ac{mAP} & Top-1 \\ \midrule
    SIFT + \ac{vlad} & 47.3 & 82.4 & 55.8 & 85.6 \\ %
    ResNet20 + \ac{vlad} & 60.9 & 90.2 & 74.9 & 94.8\\
    ResNet20 + NetVLAD & \textbf{68.6} & \textbf{94.2} & \textbf{75.2} & \textbf{95.6}\\
    \bottomrule
    \end{tabular} 
\end{table}

\begin{figure}
    \centering
        
\definecolor{darkgray176}{RGB}{176,176,176}
\definecolor{steelblue31119180}{RGB}{31,119,180}
\begin{tikzpicture}
\pgfplotsset{grid style={dashed,gray}}
       
\begin{axis}[
legend style={nodes={scale=0.75, transform shape}},
name=ax1,
width=0.48\textwidth,
tick align=inside,
tick pos=left,
xmin=0, xmax=5100,
xtick style={color=black},
xtick={1000, 2000, 3000, 5000},
xticklabels={1k, 2k, 3k, 5k},
ymin=0, ymax=80,
ytick style={color=black},
xlabel= {Number of features},
ylabel=\ac{mAP},
xmajorgrids,
ymajorgrids,
legend style={font=\small, at={(-0.1,0.-0.4)},anchor=north west,legend columns=3},
] 

\addplot [semithick, color=steelblue31119180, mark=*]
table {%
10 0.7
50 1.2
100 1.7
250 3.8
500 7.4
1000 14.7
1500 21.2
2000 27.0
3000 35.9
5000 47.3
};
\addlegendentry{\ac{sift} + \ac{vlad}}

\addplot [semithick, color=orange, mark=*]
table {%
10 3.0
50 13.2
100 22.0
250 37.2
500 47.8
1000 55.1
1500 57.8
2000 59.0
3000 60.1
5000 60.9
};
\addlegendentry{ResNet20 + \ac{vlad}}

\addplot [semithick, color=red, mark=*]
table {%
10 5.3
50 20.3
100 32.1
250 48.0
500 58.0
1000 64.1
1500 66.1
2000 67.1
3000 67.9
5000 68.5
};
\addlegendentry{ResNet20 + NetVLAD}

\legend{}

\end{axis}

\begin{axis}[
legend style={nodes={scale=0.75, transform shape}},
at={(ax1.south east)},
xshift=2cm,
width=0.48\textwidth,
tick align=inside,
tick pos=left,
xmin=0, xmax=5100,
xtick style={color=black},
xtick={1000, 2000, 3000, 5000},
xticklabels={1k, 2k, 3k, 5k},
ymin=0, ymax=80,
ytick style={color=black},
xlabel= {Number of features},
ylabel=\ac{mAP},
xmajorgrids,
ymajorgrids,
legend style={font=\small, at={(-1.2,0.-0.4)},anchor=north west,legend columns=3},
]

\addplot [semithick, color=steelblue31119180, mark=*]
table {%
10 1.5
50 2.8
100 3.9
250 6.9
500 11.6
1000 20.6
1500 27.8
2000 34.0
3000 44.0
5000 55.8
};
\addlegendentry{\ac{sift} + \ac{vlad}}

\addplot [semithick, color=orange, mark=*]
table {%
10 8.6
50 29.8
100 43.8
250 59.4
500 67.2
1000 71.9
1500 73.4
2000 74.0
3000 74.6
5000 74.9
};
\addlegendentry{ResNet20 + \ac{vlad}}

\addplot [semithick, color=red, mark=*]
table {%
10 9.8
50 29.8
100 43.3
250 59.5
500 66.5
1000 72.1
1500 73.6
2000 74.3
3000 75.0
5000 75.2
};
\addlegendentry{ResNet20 + NetVLAD}

\end{axis}
\node at (2.1,3.8) {CVL};
\node at (8.4,3.8) {IAM};

\end{tikzpicture}  
    \caption{\textbf{Line level Retrieval}: Influence of the number of features used per line for CVL and IAM dataset.}\label{fig:nr_features_line}    
\end{figure}
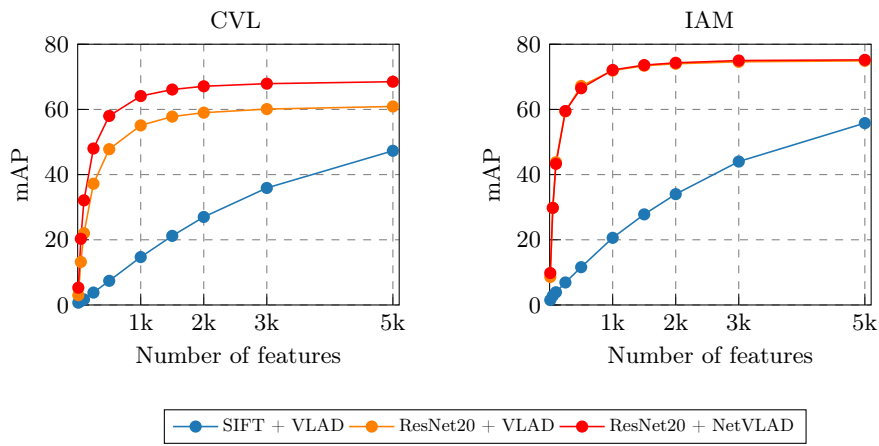

We also investigate the impact of the number of feature samples per line, which is particularly relevant for lines due to the limited amount of text. The results are presented in Fig.~\ref{fig:nr_features_line}. The most significant performance drop is observed for the traditional methods using SIFT and \ac{vlad}, where performance does not fully saturate even with 5k features per line. In contrast, for ResNet20 (\ac{vlad} and NetVLAD), the \ac{mAP} scores saturate with just 1k features per line.

\paragraph{Restricting the text quantity in the query (Short Query - Long Gallery)} Next, we focus on the retrieval task by reducing the query document’s text to a single line or half of a page and compare this to the page level performance. This setup mirrors the tasks of domain experts, such as those in forensics, who may have access to a database of reference documents but only a small handwriting sample from an unknown writer. The results are presented in Table~\ref{tab:query-exp}. For the single line and half-page experiments, we report the mean of the \ac{mAP}, as retrieval is performed multiple times (once per line and twice for each part of the page). Interestingly, we observe no significant drop in performance, indicating that retrieving relevant documents with just a line of text as the query does not negatively impact the retrieval process.

\begin{table}
\caption{mAP when using one line (1L), half the page (HP), and the full page (FP) as a query.}\label{tab:query-exp}
\centering
    \begin{tabular}{lcccccc}
    \toprule
    ~& \multicolumn{3}{c}{CVL} & \multicolumn{3}{c}{IAM} \\
    ~ & 1L & HP  & FP  &  1L & HP  & FP \\ \midrule
    SIFT + \ac{vlad} & 95.9 & 96.0 & 96.1 &  92.5 & 92.5 & 92.5 \\ 
    ResNet20 + \ac{vlad} & \textbf{97.4} & \textbf{97.4} & \textbf{97.4} & \textbf{93.7} & \textbf{93.7} & \textbf{93.8}\\
    ResNet20 + NetVLAD & 96.9 & 96.9 & 97.0 & 92.2& 92.2 &92.2 \\
    \bottomrule
    \end{tabular} 

\end{table}

\paragraph{Restricting the text quantity in the database (Short Query - Short Gallery)} As we observed that the line level \ac{wr} yields lower scores (Table~\ref{tab:wr-line}), we now aim to investigate how much text is necessary to achieve performance comparable to page level retrieval. To address this, we evaluate the methods by gradually merging the lines of a document, starting with short queries and short samples in the gallery. We begin with a single line, effectively splitting the database into its individual lines, and then progressively stack the lines per document according to the provided line order. For instance, when stacking two lines, lines 1 and 2 are combined, followed by lines 3 and 4, and so on. Remainders, or any leftover lines that do not form a complete group, are excluded from the retrieval process to ensure each sample contains exactly $n$ lines.

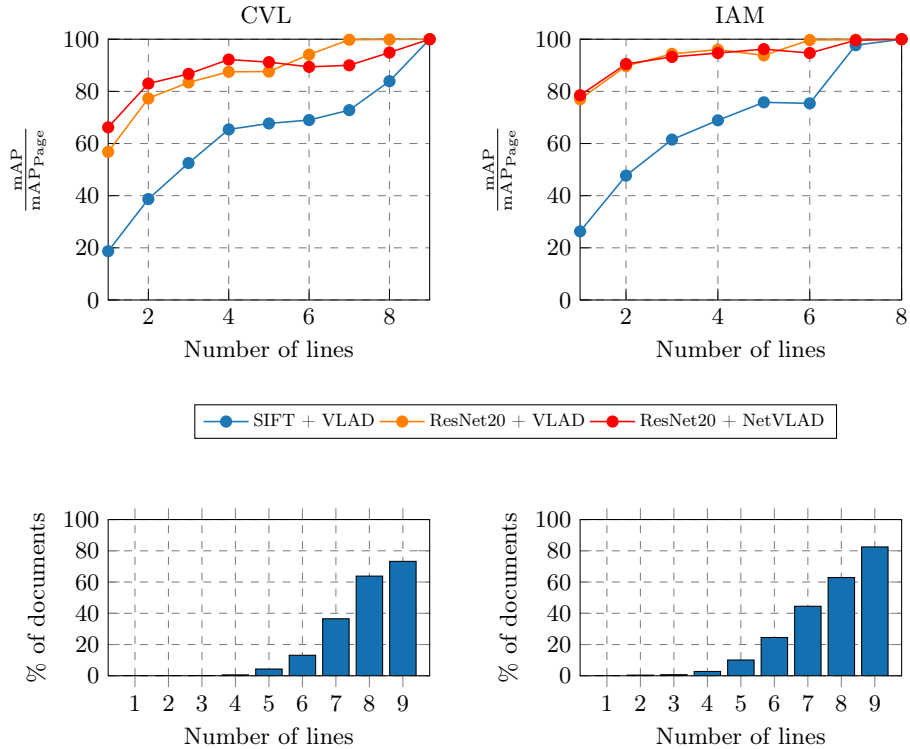
\begin{figure}
    \centering
        
\definecolor{darkgray176}{RGB}{176,176,176}
\definecolor{steelblue31119180}{RGB}{31,119,180}
\definecolor{myblue}{RGB}{20,112,176}

\begin{tikzpicture}
\pgfplotsset{grid style={dashed,gray}}
       
\begin{axis}[
legend style={nodes={scale=0.75, transform shape}},
name=ax1,
width=0.48\textwidth,
tick align=inside,
tick pos=left,
xmin=1, xmax=9,
xtick style={color=black},
ymin=0, ymax=100,
ytick style={color=black},
xlabel= {Number of lines},
ylabel=$\frac{\text{mAP}}{\text{mAP}_{\text{Page}}}$,
xmajorgrids,
ymajorgrids,
legend style={font=\small, at={(-0.1,0.-0.4)},anchor=north west,legend columns=3},
] 

\addplot [semithick, color=steelblue31119180, mark=*]
table {%
1 18.7
2 38.7
3 52.5
4 65.4
5 67.7
6 69.0
7 72.8
8 83.9
9 100
};
\addlegendentry{\ac{sift} + \ac{vlad}}

\addplot [semithick, color=orange, mark=*]
table {%
1 56.8
2 77.3
3 83.4
4 87.5
5 87.6
6 94.1
7 99.8
8 99.9
9 100
};
\addlegendentry{ResNet20 + \ac{vlad}}

\addplot [semithick, color=red, mark=*]
table {%
1 66.2
2 83.0
3 86.7
4 92.2
5 91.2
6 89.4
7 90.0
8 94.9
9 100
};
\addlegendentry{ResNet20 + NetVLAD}

\legend{}

\end{axis}

\begin{axis}[
legend style={nodes={scale=0.75, transform shape}},
at={(ax1.south east)},
xshift=2cm,
width=0.48\textwidth,
tick align=inside,
tick pos=left,
xmin=1, xmax=8,
xtick style={color=black},
ymin=0, ymax=100,
ytick style={color=black},
xlabel= {Number of lines},
ylabel=$\frac{\text{mAP}}{\text{mAP}_{\text{Page}}}$,
xmajorgrids,
ymajorgrids,
legend style={font=\small, at={(-1.2,0.-0.4)},anchor=north west,legend columns=3},
]

\addplot [semithick, color=steelblue31119180, mark=*]
table {%
1 26.3
2 47.7
3 61.5
4 68.9
5 75.8
6 75.4
7 97.7
8 100
};
\addlegendentry{\ac{sift} + \ac{vlad}}

\addplot [semithick, color=orange, mark=*]
table {%
1 77.0
2 89.7
3 94.4
4 96.0
5 93.8
6 99.7
7 99.8
8 100
};
\addlegendentry{ResNet20 + \ac{vlad}}

\addplot [semithick, color=red, mark=*]
table {%
1 78.5
2 90.5
3 93.2
4 94.7
5 96.2
6 94.7
7 99.6
8 100
};
\addlegendentry{ResNet20 + NetVLAD}

\end{axis}
\node at (2.1,3.8) {CVL};
\node at (8.4,3.8) {IAM};

\begin{axis}[
    ybar,                         
    width=0.48\textwidth,    name=ax3,
    height=0.3\textwidth,
    xlabel={Number of lines},              
    ylabel={\% of documents},               
    xtick={1, 2, 3, 4,5,6,7,8,9},           
    ymin=0,                        
xmajorgrids,
ymajorgrids,
at={(ax1.south west)},
    yshift=-5cm,
    ymax=100
]


\addplot[fill=myblue, draw=black] coordinates {(1, 0) (2, 0) (3, 0.07) (4, 0.57) (5,4.33) (6, 13.13) (7,36.5) (8,63.8) (9,73.24)};  
\end{axis}

\begin{axis}[
    ybar,                         
    at={(ax3.south east)},
    xshift=2cm,
    width=0.48\textwidth,
    height=0.3\textwidth,
    xlabel={Number of lines},              
    ylabel={\% of documents},               
    xtick={1, 2, 3, 4,5,6,7,8,9},           
    ymin=0,                        
    ymax=100,
xmajorgrids,
ymajorgrids
]

\addplot[fill=myblue] coordinates {(1, 0) (2, 0.43) (3, 0.69) (4, 2.76) (5,10.1) (6, 24.5) (7,44.5) (8,62.9) (9, 82.48)};  
\end{axis}

\end{tikzpicture}  
    \caption{\textbf{Line level Retrieval}: \ac{wr} performance normalized on the page level retrieval $\text{mAP}_\textbf{Page}$. We gradually merge the number of lines of a document for the CVL and IAM dataset.}\label{fig:merge_lines}    
\end{figure}

The results of this experiment are presented in Fig~\ref{fig:merge_lines}. In this figure, the \ac{mAP} scores are normalized relative to the page level performance reported in Table~\ref{tab:page-results}. This normalization allows for a direct comparison between the different line-merging configurations and the baseline page level retrieval. By gradually increasing the amount of text used in the retrieval, we can determine the threshold at which the performance begins to approach that of the page level retrieval, providing insight into how much textual context is necessary to achieve similar retrieval accuracy.

We observe a strong influence of the number of lines on performance, particularly for SIFT, which only achieves more than 80\% of the page level \ac{mAP} when using more than seven lines for the CVL dataset and eight lines for the IAM dataset. The deep-learning-based approaches, in contrast, perform significantly better with smaller amounts of text. However, their performance still falls below 90\% of the page level \ac{mAP} when using three or fewer lines for both datasets. 

Based on these observations, we conclude that the minimum amount of text required for performance comparable to full-page retrieval is about four lines. This threshold applies specifically to the deep-learning-based feature extractors, which reach more than 90\% of the page level performance with four lines. In contrast, the SIFT + \ac{vlad} approach performs worse, achieving only around 65\% of the page level \ac{mAP} when using four lines per sample. As the number of lines increases, the performance of all approaches improves, eventually matching the page level \ac{mAP}.

The histogram in Fig.~\ref{fig:merge_lines} illustrates the percentage of documents with fewer than $n$ lines for both datasets. We argue that the observed improvements in \ac{mAP} are due to the approaches’ ability to capture the handwriting style, rather than being attributed to the datasets being "restored." This suggests that the deep-learning methods are preferable to extract the writer-discriminate features, even for  limited text, whereas traditional methods like SIFT + \ac{vlad} require more text to reach comparable performance.

\paragraph{Line level retrieval on historical data} To show that the observations generalize on more complex handwriting, we evaluate the three WR approaches on the historical dataset Norhandv2~\cite{Tarride2024} consisting of Norwegian handwriting with a 40/60 split of the writers, since it also provides line segmentations. Firstly, the performance of the approaches is signficantly lower, as shown in Table~\ref{tab:norhandres}. However, the influence of the text quantity still holds, except that the relative drop in performance with low text is less due to the worse page level scores.

\begin{figure}
  \centering

  \begin{minipage}[t]{0.48\textwidth}
    \centering
        
\definecolor{darkgray176}{RGB}{176,176,176}
\definecolor{steelblue31119180}{RGB}{31,119,180}
\definecolor{myblue}{RGB}{20,112,176}

\begin{tikzpicture}
\pgfplotsset{grid style={dashed,gray}}
       
\begin{axis}[
legend style={nodes={scale=0.75, transform shape}},
name=ax1,
width=\textwidth,
height=0.55\textwidth,
tick align=inside,
tick pos=left,
xmin=1, xmax=10,
xtick style={color=black},
ymin=60, ymax=100,
ytick style={color=black},
xlabel= {Number of lines},
ylabel=$\frac{\text{mAP}}{\text{mAP}_{\text{Page}}}$,
xmajorgrids,
ymajorgrids,
legend style={font=\small, at={(-0.1,0.-0.4)},anchor=north west,legend columns=3},
] 

\addplot [semithick, color=steelblue31119180, mark=*]
table {%
1 73.0
2 79.2
3 83.7
4 86.9
5 89.7
6 92.2
7 94.2
8 97.8
9 98.6
10 100
};
\addlegendentry{\ac{sift} + \ac{vlad}}

\addplot [semithick, color=orange, mark=*]
table {%
1 78.3
2 82.4
3 88.4
4 91.7
5 93.8
6 95.3
7 96.8
8 97.6
9 98.5
10 99.4
};
\addlegendentry{ResNet20 + \ac{vlad}}

\addplot [semithick, color=red, mark=*]
table {%
1 84.2
2 86.2
3 89.2
4 92.2
5 94.2
6 95.6
7 96.9
8 98.5
9 99.6
10 100.0
};
\addlegendentry{ResNet20 + NetVLAD}

\legend{}
\addplot[mark=none, black, domain=1:10] {90};

\end{axis}

\end{tikzpicture}  
    \captionof{figure}{Norhandv2: Line level retrieval}
    \label{fig:norhand_lines}
  \end{minipage}
  \hfill
  \begin{minipage}[b]{0.48\textwidth}
    \centering
    \captionof{table}{Norhandv2: Full page retrieval performance}
    \label{tab:norhandres}
    \begin{tabular}{lcc}
      \toprule
      ~ & mAP & Top-1 \\
      \midrule
     SIFT + \ac{vlad} & 21.1 & 78.6 \\
     ResNet20 + \ac{vlad} & \textbf{35.6} & \textbf{92.5} \\
     ResNet20 + NetVLAD & 33.2 & 92.2\\
      \bottomrule
    \end{tabular}
  \end{minipage}
\end{figure}

\begin{table}
\caption{Hard criterion and mAP evaluated on word level with 500 features used per word.}\label{tab:wr-word}
\begin{subtable}[h]{0.95\textwidth}
\centering
    \caption{CVL}\label{tab:wr-page_cvl}
    \begin{tabular}{lccccc}
    \toprule
    ~ & Top-1 &Top-3  & Top-5  & Top-10  & mAP\\ \midrule
    SIFT + \ac{vlad} & 21.8 & 4.9 & 0.9 &  0.8 & 3.0 \\ 
    ResNet20 + \ac{vlad} & 35.0 & 12.7 & 7.1 & 3.0 & 7.5\\
    ResNet20 + NetVLAD & \textbf{49.3} & \textbf{22.8} & \textbf{13.9} &\textbf{6.2} & \textbf{9.3} \\
    \bottomrule
    \end{tabular} 
\end{subtable}

\begin{subtable}[h]{0.95\textwidth}
\centering
    \caption{IAM}
    \begin{tabular}{lccccc}
    \toprule
    ~ & Top-1 &Top-3  & Top-5  & Top-10  & mAP\\ \midrule
    SIFT + \ac{vlad} & 30.0 & 12.5 & 8.5 &  5.2 & 5.0 \\ 
    ResNet20 + \ac{vlad} & 47.2 & 27.4 & 20.5 & 13.5 & \textbf{13.6}\\
    ResNet20 + NetVLAD & \textbf{54.7} & \textbf{32.7} & \textbf{24.2} & \textbf{15.2} & 13.3 \\
    \bottomrule
    \end{tabular} 
\end{subtable}
\end{table}

\subsection{Word level Retrieval}
Finally, our paper focuses on the finest granularity, namely the words. We also provide the hard Top-$x$ criterion, since the approaches achieve only \ac{mAP} values up to 9.3/13.6\% for the CVL and IAM dataset.

Table~\ref{tab:wr-word} presents word level retrieval performance using hard criterion and mAP metrics across three approaches, each utilizing 500 features per word. In the CVL dataset, ResNet20 + NetVLAD achieves the highest scores across all metrics, outperforming both SIFT + VLAD and ResNet20 + VLAD configurations. For the IAM dataset, ResNet20 + NetVLAD delivers the best Top-$x$ performances, while ResNet20 + VLAD slightly exceeds it in mAP with 13.6\% versus 13.3\%. Overall, those results indicate that word level retrieval is more complex than retrieval at larger entities, as seen in the lower performance metrics. However, the deep-learning-based approaches (especially ResNet20 + NetVLAD) perform better than traditional feature extraction methods.

\begin{figure}
    \centering
    \includegraphics[width=0.9\linewidth]{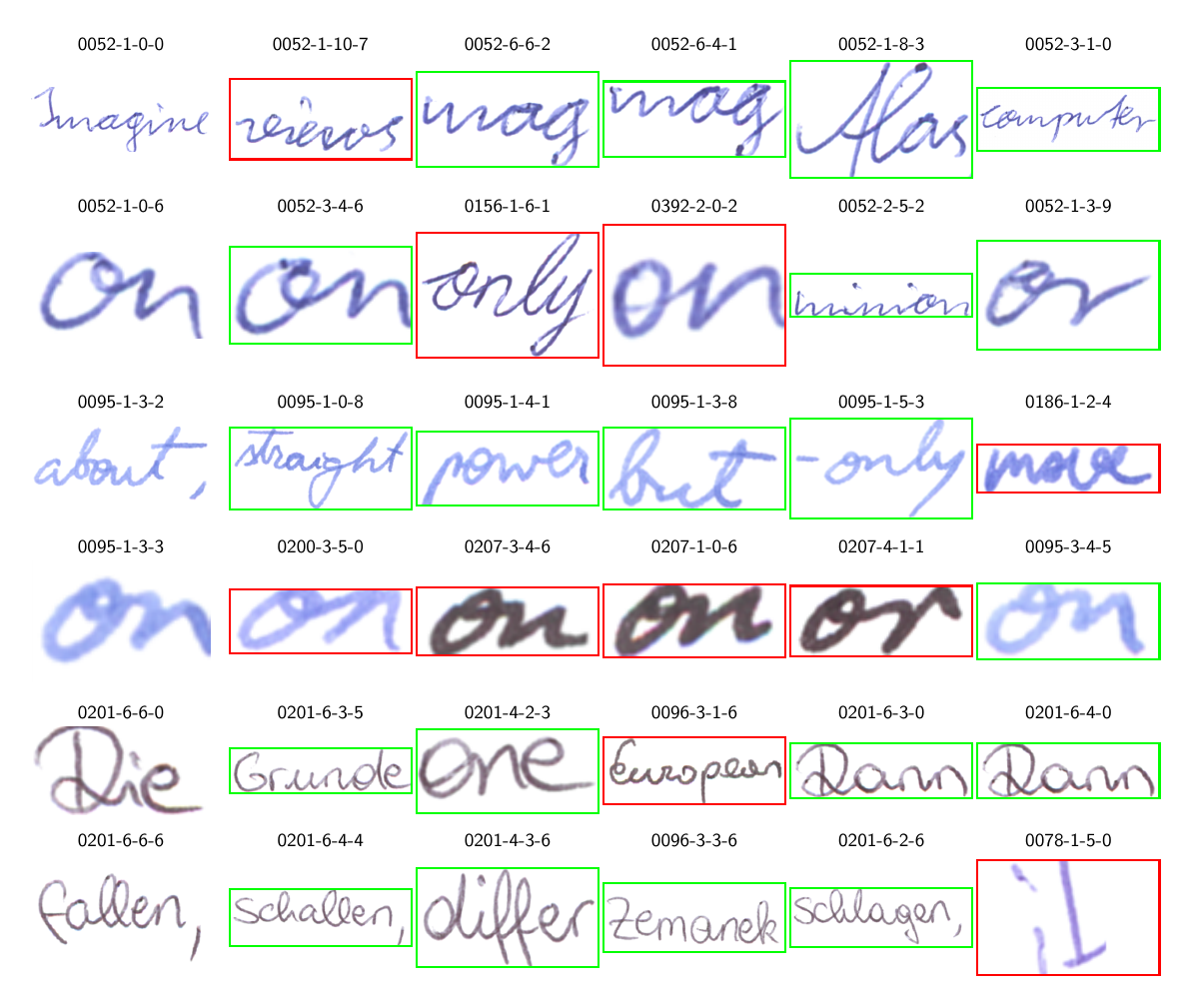}
    \caption{Qualitative results of the word retrieval using ResNet20 + NetVLAD for the CVL dataset. The left image is the query, followed by the five most similar words (IDs: \texttt{Writer-Page-Line-Word}).}
    \label{fig:qual-word-retrieval}
\end{figure}

 \paragraph{Qualitative Results} We show some qualitative results for the \ac{wr} on word level using ResNet20 with NetVLAD in Fig.~\ref{fig:qual-word-retrieval}. Firstly, the query word frequently retrieves words from the same line or adjacent lines (as seen for query \texttt{0052-1-0-0} or \texttt{0095-1-3-2}, indicating that the writing style has contextual or temporal consistency that is captured by the model. Secondly, we see that the model's results often rely on similar words, even if they are not contributed by the same writer, e.g., for \texttt{0095-1-3-3} 'on'. For other words, e.g., 'fallen' or 'Die' by 
writer \textsc{0201}, we retrieve words with similar allographs ('D', 'e' or loops of 'l', 'f').

\begin{figure}
\begin{subfigure}[h]{0.95\textwidth}
    \centering
    \definecolor{myblue}{RGB}{20,112,176}
\begin{tikzpicture}
\begin{axis}[%
    ybar,
    width=0.99\textwidth,
    height=1.8in,
    axis x line=center,
    axis y line=left,
    symbolic x coords={{of}, {the}, {a}, {on}, {and}, {in}, {or}, {Dann}, {is}, {my}, {but}, {du}, {will}, {that}, {his}, {with}, {it}, {like}, {which}, {other}},
    enlargelimits=true,
    ymin=0,
    ymax=80,
    nodes near coords,
    nodes near coords align={vertical},  
    ylabel style={align=center},
    ylabel={mAP},
    every node near coord/.append style={font=\tiny, rotate=90,anchor=east,color=white},
    xtick=data,  
    x tick label style={rotate=90,anchor=east}
]
\addplot[fill=myblue] coordinates {
    ({of}, 31.9) ({the}, 38.2) ({a}, 20.3) ({on}, 31.6) ({and}, 45.2) ({in}, 30.0) ({or}, 28.0) ({Dann}, 71.4) ({is}, 28.9) ({my}, 51.29) ({but}, 50.7) ({du}, 47.6) ({will}, 39.6) ({that}, 51.2) ({his}, 52.0)     ({with}, 52.7) ({it}, 25.6) ({like}, 49.2) ({which}, 48.2) ({other}, 35.1)
};
\end{axis}
\end{tikzpicture}
    \caption{CVL}\label{fig:word-results}    
\end{subfigure}

\begin{subfigure}[h]{0.95\textwidth}
         \centering
    \definecolor{myblue}{RGB}{20,112,176}
\begin{tikzpicture}
\begin{axis}[%
    ybar,
    width=0.99\textwidth,
    height=1.8in,
    axis x line=center,
    axis y line=left,
    symbolic x coords={{the}, {of}, {to}, {and}, {a}, {in}, {that}, {was}, {is}, {he}, {for}, {I}, {had}, {his}, {be}, {as}, {it}, {with}, {on}, {The}},
    enlargelimits=true,
    ymin=0,
    ymax=80,
    nodes near coords,
    nodes near coords align={vertical},  
    ylabel style={align=center},
    ylabel={mAP},
    every node near coord/.append style={font=\tiny, rotate=90,anchor=east,color=white},
    xtick=data,  
    x tick label style={rotate=90,anchor=east}
]

\addplot[fill=myblue] coordinates {
    ({the}, 51.9) ({of}, 49.7) ({to}, 47.5) ({and}, 60.7) ({a}, 36.2) 
    ({in}, 46.9) ({that}, 66.0) ({was}, 60.9) ({is}, 49.0) ({he}, 60.2) 
    ({for}, 58.43) ({I}, 41.2) ({had}, 70.5) ({his}, 62.9) ({be}, 58.6) 
    ({as}, 56.3) ({it}, 49.4) ({with}, 65.0) ({on}, 54.9) ({The}, 59.3)
};
\end{axis}
\end{tikzpicture}
         \caption{IAM}
         \label{fig:three sin x}
     \end{subfigure}
\caption{\ac{wr} performance for the 20 most common words in the respective dataset.}\label{fig:wwr}    
\end{figure}
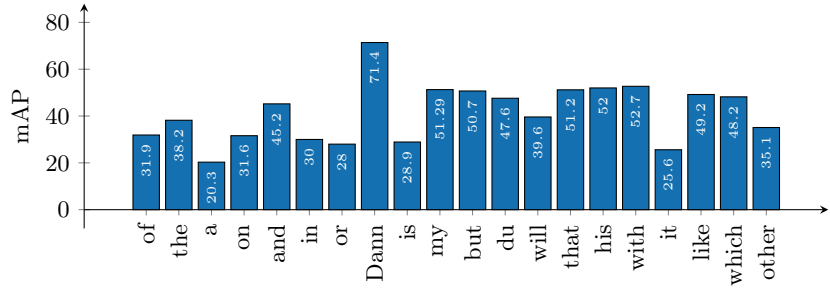
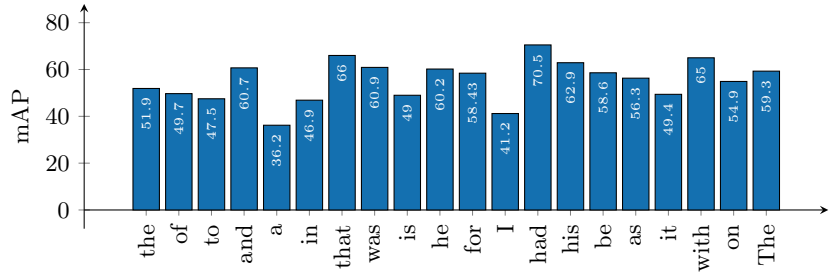

\paragraph{Word-Specific Retrieval} Lastly, we introduce the task of \emph{word-specific retrieval}, where the database consists solely of samples containing a specific word, and the goal is to retrieve samples written by the same writer. We focus on evaluating the ResNet20 combined with NetVLAD, as it provides the best performance. In Fig.~\ref{fig:wwr}, we show the \ac{mAP} performances for the 20 most common words in both datasets. Compared to full word-based \ac{wr}, the results improve for two main reasons: 1) the reduced number of relevant items (limiting the samples per writer to just one word), and 2) the ability to compare allographs of specific characters within the word. For instance, even for the single character 'a', we achieve \ac{mAP} scores of 29.3\% and 36.2\%. On the CVL dataset, the German word 'Dann' (then) achieves the highest \ac{mAP} of 71.4\%, likely due to its repeated occurrences on the same page. However, we observe no clear pattern suggesting that words with more letters yield better performance. We conclude that retrieval based on identical word instances is more advantageous for current \ac{wr} methods—especially when the available handwriting is limited—even when the methods are trained in a text-independent manner.

\section{Conclusion}\label{sec:conclusion}
In this paper, we introduce new benchmarks for three state-of-the-art approaches in the domain of handwriting retrieval (\ac{wr}) on contemporary handwriting, specifically using the CVL and IAM datasets. We present results at the page, line, and word levels, analyzing the impact of feature sample size for the global descriptor and the amount of text, with a particular focus on line retrieval. Our findings indicate that while the methods achieve over 95\% \ac{mAP} at the page level, their performance drops by approximately a third when only one line of text is available. Additionally, our experiments reveal that retrieval performance improves as the number of lines increases, with performance approaching 90\% of page level retrieval accuracy for four lines of text. On word level retrieval, however, current methods perform poorly. We also demonstrate that writer-specific word retrieval, where the database contains only specific words, yields better results than databases with multiple, different words.

For future work, we suggest that \ac{wr} methods be evaluated in low-text scenarios to gain a clearer understanding of current algorithms' limitations. We also propose that text-dependent methods may prove more effective when the available handwriting per writer is limited, such as in the case of historical documents, where comparing identical word instances leads to improved performance. Finally, we recommend exploring line- or word level retrieval to facilitate the development of more advanced feature aggregation methods (e.g., learned variants instead of simple sum pooling), as aggregation of features may be more impactful when the available feature set is small.

%
\bibliographystyle{splncs04}
\bibliography{bib,cvl_tr}

\end{document}